\documentclass[nojss,shortnames]{jss} 

\usepackage{amssymb}

\usepackage{orcidlink,thumbpdf,lmodern}

\usepackage{framed}
\usepackage{fancyvrb}

\usepackage{tikz}
\usepackage{caption}
\usetikzlibrary{shapes,arrows,positioning}

\author{
Connor T. Jerzak~\orcidlink{0000-0003-1914-8905}
\\ The University of Texas at Austin
\\  \url{ConnorJerzak.com}
   \And 
Adel Daoud~\orcidlink{0000-0001-7478-8345}
\\  Link\"oping University
\\ \url{AdelDaoud.se}
}
\Plainauthor{Connor T. Jerzak, Adel Daoud}

\title{ CausalImages: An R Package for Causal Inference with Earth Observation, Bio-medical, and Social Science Images}
\Plaintitle{ CausalImages: An R Package for Causal Inference with Earth Observation, Bio-medical, and Social Science Images }
\Shorttitle{CausalImages: A Package for Causal Inference with Images}

\Abstract{
The {\bf causalimages} R package enables causal inference with image and image sequence data, providing new tools for integrating novel data sources like satellite and bio-medical imagery into the study of cause and effect. One set of functions enables image-based causal inference analyses. For example, one key function decomposes treatment effect heterogeneity by images using an interpretable Bayesian framework. This allows for determining which types of images or image sequences are most responsive to interventions. A second modeling function allows researchers to control for confounding using images. The package also allows investigators to produce embeddings that serve as vector summaries of the image or video content. Finally, infrastructural functions are also provided, such as tools for writing large-scale image and image sequence data as sequentialized byte strings for more rapid image analysis. {\bf causalimages} therefore opens new capabilities for causal inference in {\bf R}, letting researchers use informative imagery in substantive analyses in a fast and accessible manner. 
\\ \vspace{0.2cm}
{\sc \noindent Repository:} {\sf GitHub.com/AIandGlobalDevelopmentLab/causalimages-software}
}

\Plainkeywords{Causal inference, image data, image sequence data, computer vision, machine learning, R}

\Address{
Connor T. Jerzak\\
  Department of Government\\
University of Texas at Austin \\
110 Inner Campus Drive \\ 
 Austin, TX 78712 \\ 
  E-mail: \email{connor.jerzak@austin.utexas.edu}\\
  URL: \url{ConnorJerzak.com}
\\ \\ 
Adel Daoud\\
Institute for Analytical Sociology\\
Link\"oping University \\
SE-581 83 Link\"oping
Sweden \\ 
  E-mail: \email{adel.daoud@liu.se}\\
  URL: \url{AdelDaoud.se}
}

\usepackage{graphicx}
\graphicspath{{./figures/}}
\usepackage{float} 

\begin{document}


\section{Introduction: Causal Inference with Images}\label{s:Intro}

\begin{figure}[H] \centering
\includegraphics[width=0.75\linewidth]{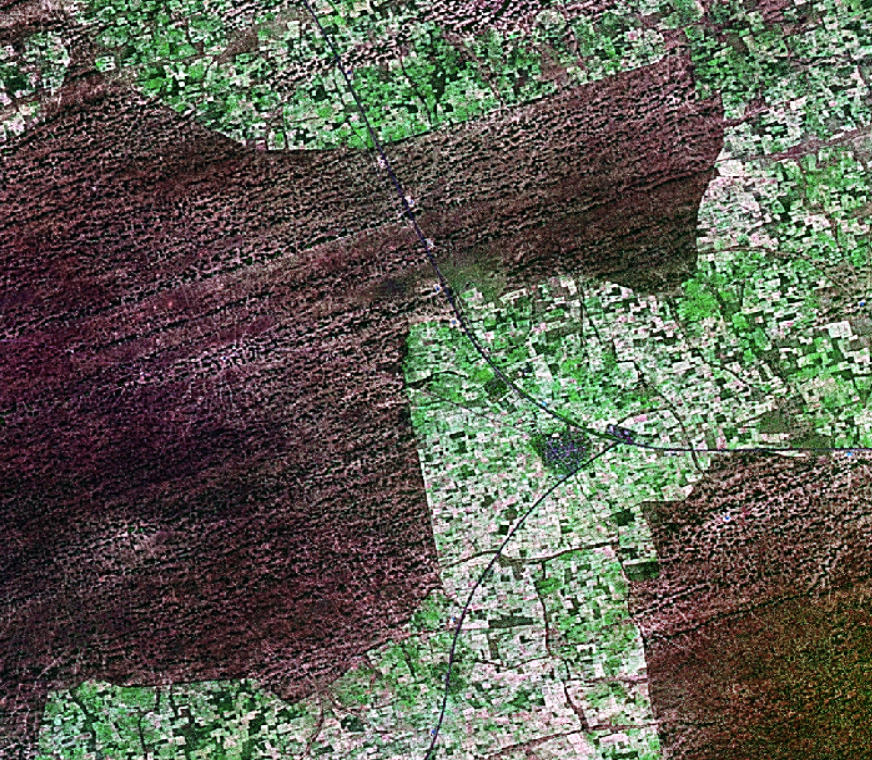}
\caption{ Landsat satellite imagery over Nigeria. }  \label{f:NigeriaViz}
\end{figure}

Satellite image data represents an emerging resource for research in global development and earth observation, yet no R package currently exists to handle images for causal inference up to now. By \textit{causal inference}, we refer to the rich literature in statistics \citep{imbens2016causal}, computer science \citep{pearl_causality_2009}, and beyond \citep{hernan_causal_2020}.  Satellites generate temporally-rich worldwide coverage, capturing the entire Earth's surface at regular intervals, except when obscured by clouds \citep{burke_using_2021}. Historical archives date back to the 1970s. Unlike snapshots of political, economic, or educational systems at a single time point, satellites revisit each location every 2 weeks or more, providing approximately 26 temporal observations annually. This time-series information has proven valuable for studying phenomena like transportation network growth \citep{nagne2013transportation}, urbanization \citep{schneider2009new}, health and living conditions \citep{daoud_using_2023,chi_microestimates_2022}, living standards \citep{yeh2020using,ijcai2023p684}, and neighborhood characteristics \citep{sowmya2000modelling}. Thus, satellite data facilitates observational inference where ground-level data is lacking. Moreover, image quality and frequency continue improving as the satellite population proliferates from hundreds to thousands \citep{tatem2008fifty}, with sub-100 cm resolution now available \citep{hallas2019mapping}. We see an example of this data source in Figure~\ref{f:NigeriaViz}.

Methodological guidance has remained limited for causal estimation from satellite images \citep{daoud_statistical_2023}. To address that methodological gap, recently \citet{jerzak2022estimating} proposed methods to estimate confounding, and \citet{jejoda2022_hetero}  developed methods for estimating effect heterogeneity in images. Our {\bf causalimages} package encompasses these methods and some future ones. Our confounding method helps address that research need by examining observational causal inference amidst image-based confounding. Our heterogeneity method shows how researchers may use past satellite images to proxy geographical and historical processes important for moderating the treatment effect of both randomized experiments and observational studies. 

Our causal inference work with images complements the growing use of visual data in climate science, sociology, economics, political science, and biomedical research \citep{kino_scoping_2021,daoud_statistical_2023}. Examples include qualitative photo analysis \citep{pauwels_visual_2010,ohara_participant_2019}, image similarity calculation \citep{zhang_image_2022}, crowd size estimation \citep{cruz_unbiased_2021}, and relating social outcomes to Street View scenes \citep{gebru2017using}. Recent extensions encompass video for investigating social processes like police violence \citep{nassauer_video_2021}. In large-data quantitative studies, algorithms have been trained to identify objects of interest automatically \citep{torres_learning_2022}. Similarly, in the biomedical domain, \citet{castroCausalityMattersMedical2020} shows how a variety of image data---from X-ray to ultrasound pictures to MRI scans---can be used for causal inference. However, more research is needed to close the gap between foundational and applied research across those domains, and not only earth observation data. To contribute to closing that gap, we created  {\bf causalimages}. Although our examples focus on satellite images, research can use any image data and across other domains where image data are available.

In concluding this introduction, we note that this package builds on the \proglang{R} ecosystem regarding data visualization and geospatial analysis. For instance, the \proglang{R} packages like \pkg{tensorflow} and \pkg{keras} provide the backbone 
for deep learning functionalities with the TensorFlow backend. The \pkg{viridis} package enhances the 
visualization capabilities, while \pkg{animation} facilitates dynamic plots for results involving image sequence data. Integration with 
Python is streamlined through \pkg{reticulate}. For geospatial operations, we rely on \pkg{geosphere} 
and \pkg{raster}. 


\section{Models and Software} \label{s:Software}

\subsection{Package Overview}
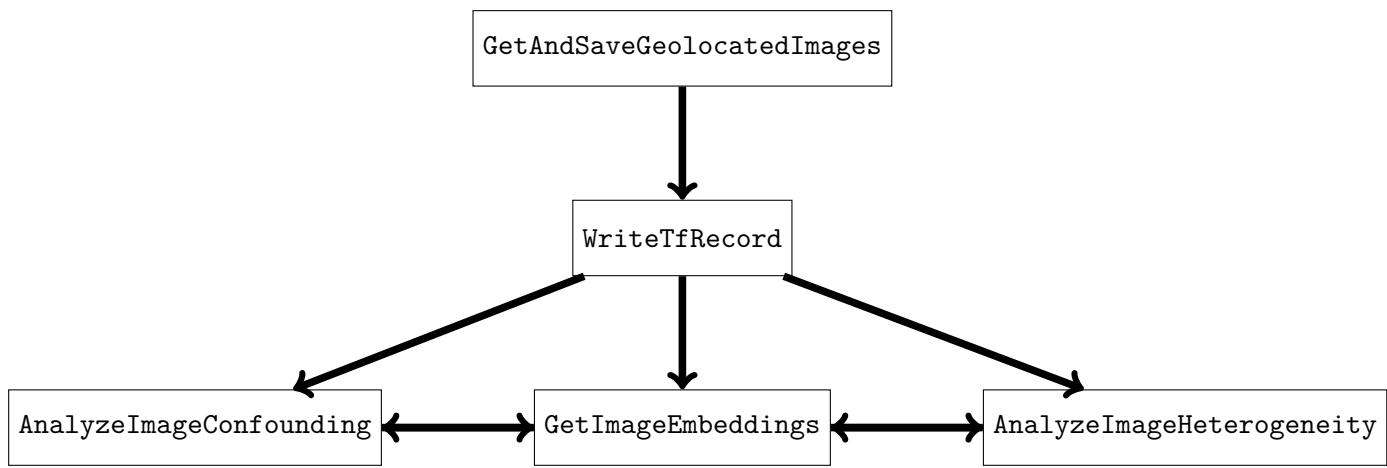
\begin{figure}[H]
\centering
\begin{tikzpicture}[
    node distance=1.5cm and 2cm,
    box/.style={rectangle, draw, align=center, minimum width=1cm, minimum height=1cm}
]

\node[box] (getimages) {\texttt{GetAndSaveGeolocatedImages}};
\node[box, below=of getimages] (writetfrecord) {\texttt{WriteTfRecord}};
\node[box, below=of writetfrecord] (getembeddings) {\texttt{GetImageEmbeddings}};
\node[box, left=of getembeddings] (analyzeconfounding) {\texttt{AnalyzeImageConfounding}};
\node[box, right=of getembeddings] (analyzeheterogeneity) {\texttt{AnalyzeImageHeterogeneity}};

\draw[->, line width=1mm] (getimages) -- (writetfrecord);
\draw[->, line width=1mm] (writetfrecord) -- (getembeddings);
\draw[->, line width=1mm] (writetfrecord) -- (analyzeheterogeneity);
\draw[->, line width=1mm] (writetfrecord) -- (analyzeconfounding);
\draw[->, line width=1mm] (getembeddings) -- (analyzeheterogeneity);
\draw[->, line width=1mm] (getembeddings) -- (analyzeconfounding);
\draw[->, line width=1mm] (analyzeheterogeneity) -- (getembeddings);
\draw[->, line width=1mm] (analyzeconfounding) -- (getembeddings);

\end{tikzpicture}
\caption{Overview of workflow in the \texttt{causalimages} package.}\label{f:FGraph}
\end{figure}

At a high level, the \texttt{causalimages} package provides tools for causal inference with image data. The package contains several functions whose relations are summarized in Figure \ref{f:FGraph}.

The \texttt{AnalyzeImageConfounding} and \texttt{AnalyzeImageHeterogeneity}, functions run the main analysis models for image causality. They require observed treatment and outcome data, as well as a way to retrieve image data associated with each observation.

Other functions center on working with image data in a more infrastructural sense. The \texttt{GetAndSaveGeolocatedImages} function helps retrieve image data referenced by geographic coordinates. \texttt{WriteTfRecord} writes image or image sequence data to a sequentialized TFRecord file for efficient retrieval. \texttt{GetImageEmbeddings} generates embeddings useful for other tasks where an efficient summary of the image information is required. 

Together, these tools enable causal analyses that incorporate image data as confounders, mediators, and moderators of treatment effects. The analyses can adjust for spatial dependencies and estimate heterogeneous treatment effects associated with geospatial imagery.

\subsection{Package Installation and Loading}\label{s:Install}

The {\bf causalimages} package is currently installed using the \texttt{devtools} package. For  installation, users should run:
\begin{Code}
R> devtools::install_github(repo = "AIandGlobalDevelopmentLab/causalimages-software")
\end{Code}
To load the package into a live \proglang{R} environment, use: 
\begin{CodeChunk}
\begin{CodeInput}
R> library(causalimages)
\end{CodeInput}
\end{CodeChunk}

The {\bf causalimages} package uses a {\bf tensorflow} backend for image analysis and GPU utilization. Python version 3 or above is assumed. To install {\bf tensorflow} into your default Python environment, try
\begin{CodeChunk}
\begin{CodeInput}
R> library(reticulate)
R> py_install("tensorflow")
\end{CodeInput}
\end{CodeChunk}
You may need to do the same to install dependencies such as {\bf tensorflow-probability} and {\bf gc}. For more fine-grained user control over CPU and GPU use, we recommend installing the requisite backend into a conda environment.

\subsection{Tutorial Data}
Once the package has been successfully initiated, users can access package data useful for tutorial purposes. The data are drawn from an anti-poverty experiment in Uganda \citet{blattman2020long} and contain information on the treatment, experimental outcome, approximate coordinates for each unit, as well as pre-treatment covariates and geo-referenced satellite images for each unit. To allow researchers to load all images into memory, we have cropped these images to a smaller-than-original size. 
\begin{CodeChunk}
\begin{CodeInput}
R> data(CausalImagesTutorialData)
\end{CodeInput}
\end{CodeChunk}
The Blattman data are then structured as follows: 
\begin{CodeChunk}
\begin{CodeInput}
R> summary(obsW)  % A vector portraying treatment information
R> summary(obsY)  % A vector with an experimental outcome 
R> summary(LongLat)  % Geospatial coordinates for each observational unit
R> summary(X)  % Pre-treatment covariate information
R> causalimages::image2(FullImageArray[1,,,1]) # image associated with unit 1
R> causalimages::image2(FullImageArray[3,,,2]) # image associated with unit 3
\end{CodeInput}
\end{CodeChunk}

\subsection{Functions for Data Assimilation}\label{s:DataAssim}
We have several functions for helping users save geo-located images. For example, {\bf GetAndSaveGeolocatedImages} finds the image slice associated with given longitude and latitude values and saves images by band, given a pool of .tif's. For example, the .tif pool may contain dozens of large Landsat mosaics covering the continent of Africa, and we want to extract a 500$\times$500 meter square image around a particular point.  For the Blattman data, we have provided the Landsat images in the package. For other data, the user has to download data from USGS or Google Earth Engine. 

In the following example, we have two .tif's saved in \verb|"./LargeTifs"|, we can search across those images for matches to the associated \Verb|long| and \Verb|lat| inputs. When a match is found, a series of .csv's are written to encompass the data in a \verb|image_pixel_width| square around the target geo-point in the target .tif. These image objects are saved in the \verb|save_folder| as 
\verb|Key[key]_BAND[band].csv| where \Verb|[key]| refers to the appropriate entry from \Verb|keys| specifying the label for each image and \Verb|[band]| specifies the band. 
\begin{CodeChunk}
\begin{CodeInput}
R> MASTER_IMAGE_POOL_FULL_DIR <- c("./LargeTifs/tif1.tif","./LargeTifs/tif2.tif")
R> GetAndSaveGeolocatedImages(
+    long = GeoKeyMat$geo_long,
+    lat = GeoKeyMat$geo_lat,
+    image_pixel_width = 500L,
+    keys = row.names(GeoKeyMat),
+    tif_pool = MASTER_IMAGE_POOL_FULL_DIR,
+    save_folder = "./Data/Uganda2000_processed",
+    save_as = "csv",
+    lyrs = NULL)
\end{CodeInput}
\end{CodeChunk}

\subsection{Functional Image Loading}\label{s:AcquireImage}
One important part of the image analysis pipeline is writing a function that acquires the appropriate image data for each observation. This function will be fed into the \texttt{acquireImageFxn} argument of the package functions unless an approach using TFRecords is used instead. There are two ways that you can approach this: (1) you may store all images in \texttt{R}'s memory (feasible only for problems involving few or small images), or you may (2) save images on your hard drive (e.g., using {\bf GetAndSaveGeolocatedImages}) and read them in when needed. The second option will be more common for large images.

You will write your \texttt{acquireImageFxn} to take in one main argument---\texttt{keys}. \texttt{keys} is fed a character or numeric vector. Each value of \texttt{keys} refers to a unique image object that will be read in. If each observation has a unique image associated with it, perhaps \texttt{imageKeysOfUnits = 1:nObs}. If multiple observations map to the same image, then multiple observations will map to the same \texttt{keys} value. The \texttt{keys} thus serves as an identifier for a particular image, which is then referenced back to individuals who are matched with keys of their associated image.  

In practice, users should ensure that \texttt{acquireImageFxn} returns arrays with dimensions batch by height by width by channels in the case of images and batch by time by height by width by channels in the case of image sequences/videos. 

\subsubsection{When Loading All Images in Memory}
We here provide an example of writing an \texttt{acquireImageFxn} function using the tutorial data wherein the images are already read into memory. 
\begin{CodeChunk}
\begin{CodeInput}
R> acquireImageFromMemory <- function(keys, training = F){
+  m_ <- FullImageArray[match(keys, KeysOfImages),,,]
+  if(length(keys) == 1){
+    m_ <- array(m_,dim = c(1L,dim(m_)[1],dim(m_)[2],dim(m_)[3]))
+  }
+  return( m_ )
+}
\end{CodeInput}
\end{CodeChunk}

To run this \texttt{acquireImageFunction} in practice, we would take 
\begin{CodeChunk}
\begin{CodeInput}
R> ImageSet <- acquireImageFromMemory(KeysOfObservations[c(5,7)])
\end{CodeInput}
\end{CodeChunk}
where ImageSet contains the images associated with observations 5 and 7 (note that they could both have the same image if these units are co-located). 

\subsubsection{When Reading in Images from Disk}
For most applications of large-scale causal image analysis, we won't be able to read the whole set of images into \texttt{R}'s memory. Instead, we can specify a function that will read images from somewhere on your hard drive. You can also experiment with other methods---as long as you can specify a function that returns an image when given the appropriate \texttt{imageKeysOfUnits} value, you should be fine. Here's an example of an \texttt{acquireImageFxn} that reads images from disk:

\begin{CodeChunk}
\begin{CodeInput}
R> acquireImageFromDisk <- function(keys, training = F){
+  array_shell <- array(NA,dim = c(1L,imageHeight,imageWidth,NBANDS))
+  array_ <- sapply(keys,function(key_){
+    for(band_ in 1:NBANDS){
+      array_shell[,,,band_] <-
+        (as.matrix(data.table::fread(
+          input = sprintf("./Data/Uganda2000_processed/Key%s_BAND%s.csv",
                          key_,
                          band_),header = F)[-1,] ))
+    }
+    return( array_shell )
+  }, simplify="array")
+  array_ <- tf$squeeze(tf$constant(array_,dtype=tf$float32),0L)
+  array_ <- tf$transpose(array_,c(3L,0L,1L,2L))
+  return( array_ )
+}
\end{CodeInput}
\end{CodeChunk}

\subsection{TFRecords Integration for Fast Image Processing}
We can use the aforementioned functions for acquiring images from disk to write the data corpus in an optimized format for fast reading-writing. This format is not required but is highly recommended to improve causal image analysis runtimes. 

In particular, once we have acquired a pool of geo-referenced satellite images, {\bf causalimages} also contains a function that writes the analysis data in TFRecord format, a binary storage format used by TensorFlow to store data efficiently and to enable fast acquisition of data into memory in serialized chunking---a process that speeds up the acquisition of images into memory where data are too large to fit into memory. 
\begin{CodeChunk}
\begin{CodeInput}
R> WriteTfRecord(
+    file = "./UgandaApp.tfrecord",
+    keys = KeysOfObservations,
+    acquireImageFxn = acquireImageFromMemory,
+    conda_env = "tensorflow_m1")
\end{CodeInput}
\end{CodeChunk}
{\bf WriteTfRecord} writes to TFRecords format the entire image data stream. As we discuss later, this same function can be used if the inputted \texttt{acquireImageFxn} function outputs image sequences. 

\subsection{Image and Image Sequence Embeddings}\label{s:Embeddings}
The \texttt{GetImageEmbeddings} function offers a methodology for the extraction of image and video embeddings, particularly tailored for earth observation tasks that drive causal inference. Using the randomized convolutions approach in  \citet{rolf2021generalizable}, the function generates vector representations of images and image sequences based on the similarity within these data to a large set of smaller image patterns (i.e. kernels). The parameters provided to the function allow fine-tuned control over the embedding process, especially in the kind of convolutional kernels used. This flexibility can allow the embeddings to be adapted for the given dataset. 

We note that the embeddings function works with both image and image sequence data. It can also be run in the de-confounding and heterogeneity decomposition functions we will analyze later. 

To use the function, you can specify how to load images via the \texttt{acquireImageFxn}
\begin{CodeChunk}
\begin{CodeInput}
R> MyImageEmbeddings <- GetImageEmbeddings(
+    imageKeysOfUnits = KeysOfObservations[ take_indices ],
+    acquireImageFxn = acquireImageFromMemory,
+    nEmbedDim = 100,
+    kernelSize = 3L,
+    conda_env = "tensorflow_m1",
+    conda_env_required = T
)
\end{CodeInput}
\end{CodeChunk}
Again, \verb|conda_env| specifies a conda environment in which the desired version of the TensorFlow backend lives. If \texttt{NULL}, we search in the default Python environment for the backend. 

Alternatively, you may use the \texttt{tfrecords} approach as follows: 
\begin{CodeChunk}
\begin{CodeInput}
R> MyImageEmbeddings <- GetImageEmbeddings(
+    file = "./UgandaApp.tfrecord",
+    nEmbedDim = 100,
+    kernelSize = 3L,
+    conda_env = "tensorflow_m1",
+    conda_env_required = T
)
\end{CodeInput}
\end{CodeChunk}

Finally, we can also obtain embeddings over image sequences. To do so, we first write a simple function creating image sequences given \verb|keys|.
\begin{CodeChunk}
\begin{CodeInput}
R> acquireVideoRepFromMemory <- function(keys, training = F){
+    tmp <- acquireImageFromMemory(keys, training = training)
+ 
+    if(length(keys) == 1){
+        tmp <- array(tmp,dim = c(1L,dim(tmp)[1],dim(tmp)[2],dim(tmp)[3]))
+    }
+ 
+    tmp <- array(tmp,dim = c(dim(tmp)[1],
+                             2,
+                             dim(tmp)[3],
+                             dim(tmp)[4],
+                             1L))
+    return(  tmp  )
+}
\end{CodeInput}
\end{CodeChunk}
To obtain video embeddings, we  take: 
\begin{CodeChunk}
\begin{CodeInput}
R> MyVideoEmbeddings <- GetImageEmbeddings(
+    imageKeysOfUnits = KeysOfObservations[ take_indices ],
+    acquireImageFxn = acquireVideoRepFromMemory,
+    temporalKernelSize = 2L,
+    kernelSize = 3L,
+    nEmbedDim = 100,
+    conda_env = "tensorflow_m1",
+    conda_env_required = T)
\end{CodeInput}
\end{CodeChunk}
We can also write a TFRecord and use that in obtaining image sequence embeddings by specifying the \texttt{file} argument. 

\subsection{Deconfounding with Image and Image Sequence}\label{s:Confound}
Using the {\bf AnalyzeImageConfounding} function, causal effects are estimated using image-based or image-sequence-based confounders, although we add the option to include tabular confounders as well.

\begin{CodeChunk}
\begin{CodeInput}
R> ImageConfoundingAnalysis <- AnalyzeImageConfounding(
+    obsW = obsW[ take_indices ],
+    obsY = obsY[ take_indices ],
+    X = X[ take_indices,apply(X[ take_indices,],2,sd)>0],
+    long = LongLat$geo_long[ take_indices ],
+    lat = LongLat$geo_lat[ take_indices ],
+    imageKeysOfUnits = KeysOfObservations[ take_indices ],
+    acquireImageFxn = acquireImageFromMemory,
+    batchSize = 4,
+    #modelClass = "cnn", # uses convolutional network (richer model class)
+    modelClass = "embeddings", # uses image embeddings (faster)
+    file = NULL,
+    plotBands = c(1,2,3),
+    dropoutRate = 0.1,
+    tagInFigures = T, figuresTag = "TutorialExample",
+    nBoot = 10,
+    nSGD = 10, # this should be more like 1000 in full analysis
+    figuresPath = "~/Downloads", # figures saved here
+    conda_env = "tensorflow_m1", 
+    conda_env_required = T
+)
\end{CodeInput}
\end{CodeChunk}
{\bf AnalyzeImageConfounding} returns a list containing an image-adjusted ATE estimate \verb|tauHat_propensityHajek| (see \cite{jerzak2022estimating} for details) and an uncertainty estimate,
\verb|tauHat_propensityHajek_se|.
A matrix of out-of-sample performance metrics (e.g., out-of-sample negative log-likelihood) is housed in \verb|ModelEvaluationMetrics|. Users can specify whether they would like to use the faster option, \verb|modelClass = "embeddings"|, or the more computationally intensive but richer modeling approach using end-to-end convolutional neural network (CNN) training \verb|modelClass = "cnn"|. To speed up performance, we recommend letting \verb| acquireImageFxn = NULL| and instead writing and then specifying a TFRecords file  (e.g., set \verb|file| to the path of the TFRecord saved via a call to \verb|WriteTfRecord|).

\begin{figure}[H]
 \centering \includegraphics[width=1.\textwidth]{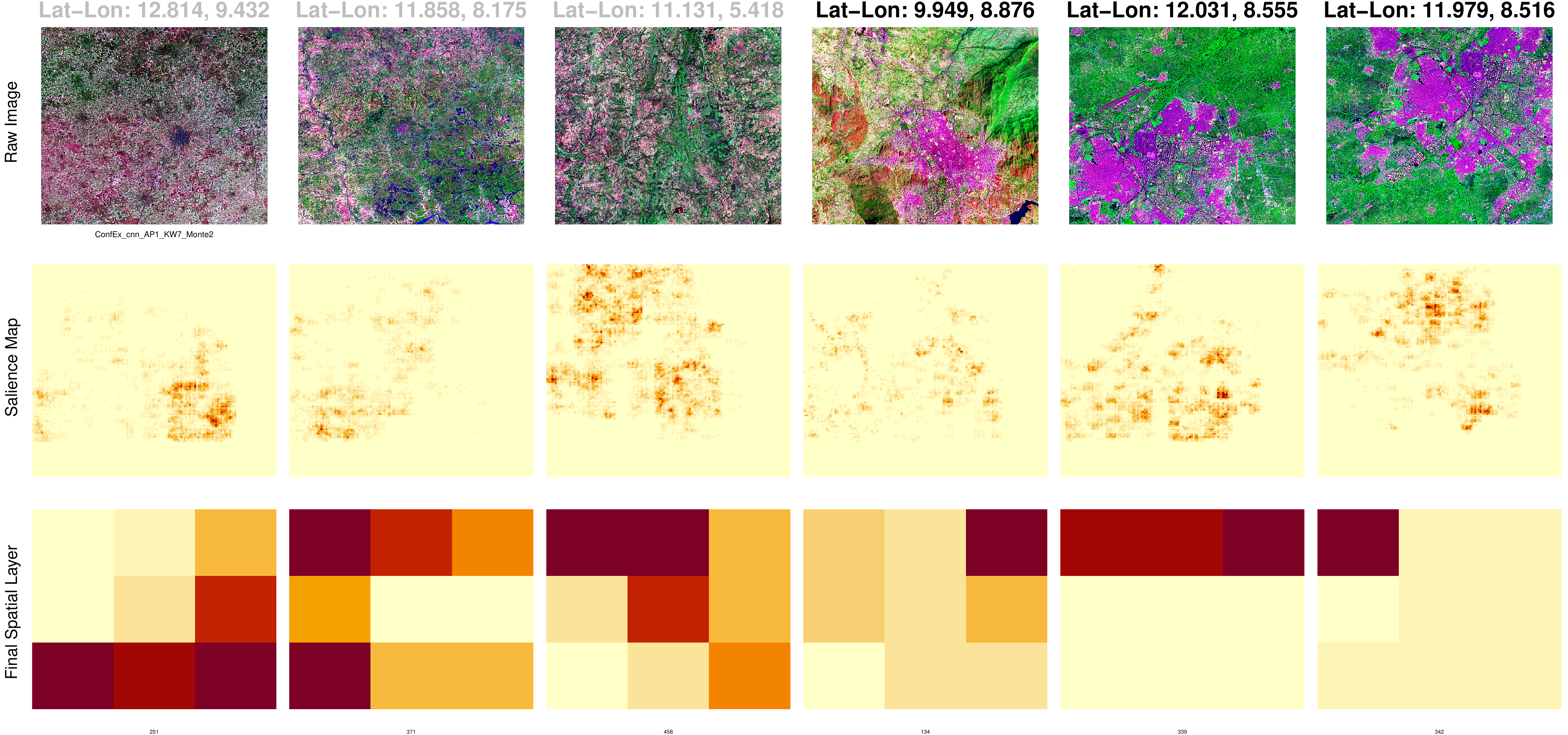}
    \caption{The three panels on the left illustrate the unprocessed data for control units, relevance diagrams for the predicted treatment likelihood, and the outcome from the concluding spatially detailed layer of the CNN image framework. The trio of panels on the right shows comparable elements for the treated units.}\label{fig:InsideConvnet}
\end{figure}

In addition to providing these estimated quantities, the function also writes to disk summary PDF outputs containing salience maps, which quantify areas in the image or image sequence that, if changed, would lead to the largest change in predicted treatment probability. An example of such figures is found in Figure \ref{fig:InsideConvnet}. Note that this figure can be made for either the embeddings or CNN image modeling backbone. 

\subsection{Heterogeneity Analysis with Image and Image Sequence Data }\label{s:Hetero}
Using the {\bf AnalyzeImageHeterogeneity} function, Conditional Average Treatment Effects (CATEs) are estimated using image-based or image-sequence-based pre-treatment information, although we add the option to include tabular pre-treatment covariates as well if the \texttt{X} argument is fed a numeric matrix input. This function would be used, for example, if we would want to learn about the kinds of geographies or developmental trajectories, proxied by satellite image data, most conducive to favorable responses to anti-poverty interventions. There are also possible applications in the biomedical domain where image data could be associated with a high or low response to a drug. 

The functionality of {\bf AnalyzeImageHeterogeneity} works much like {\bf AnalyzeImageConfounding}:

\begin{verbatim}
R> ImageHeterogeneityResults <- AnalyzeImageHeterogeneity(
+    # data inputs
+    obsW =  UgandaDataProcessed$Wobs,
+    obsY = UgandaDataProcessed$Yobs,
+    imageKeysOfUnits =  UgandaDataProcessed$geo_long_lat_key,
+    acquireImageFxn = acquireImageFromDisk,
+    conda_env = "tensorflow_m1", # change to your conda env
+    conda_env_required = T,
+    X = X,
+    plotBands = 1L,
+    lat =  UgandaDataProcessed$geo_lat, # not required, deals with redundant locs
+    long =  UgandaDataProcessed$geo_long, # not required, deals with redundant locs
+    
+    # inputs to control where visual results are saved as PDF or PNGs
+    plotResults = T,
+    figuresPath = "~/Downloads/",
+    printDiagnostics = T,
+    figuresTag = "causalimagesTutorial",
+    
+    # optional arguments for generating transportability maps
+    transportabilityMat = NULL,
+    
+    # other modeling options
+    modelClass = "embeddings", # uses image/video embeddings model class
+    orthogonalize = F,
+    heterogeneityModelType = "variational_minimal",
+    kClust_est = 2,
+    nMonte_variational = 2L,
+    nSGD = 4L,
+    batchSize = 34L,
+    kernelSize = 3L, maxPoolSize = 2L, strides = 2L,
+    nDepthHidden_conv = 2L,
+    nFilters = 64L,
+    nDepthHidden_dense = 0L,
+    nDenseWidth = 32L,
+    nDimLowerDimConv = 3L
+)
\end{verbatim}
The function performs the image heterogeneity decomposition analysis delineated in \citet{jejoda2022_hetero}. Users specify the treatment variable and outcome data, alongside a specific function that guides the loading of images using reference keys for each unit, the function produces several outputs. All outputs are saved in a list. The \texttt{clusterTaus\_mean} serves to present the estimated image effect cluster means, while the \texttt{clusterTaus\_sd} contains the estimated standard deviations of those effect clusters. 

The function further produces the \texttt{clusterProbs\_mean}, which contains the average probabilities of these image effect clusters. Additionally, it offers an estimation of the standard deviations for these cluster probabilities through the \texttt{clusterTaus\_sd}. For a deeper dive into the probabilistic insights, the \texttt{clusterProbs\_lowerConf} gives the estimated lower confidence bounds for the effect cluster probabilities. 

Regarding treatment effects, the function computes the \texttt{impliedATE}, a derived average treatment effect, and the \texttt{individualTau\_est}, which highlights estimated treatment effects on an individual image basis. The \texttt{transportabilityMat} avails a matrix filled with cluster information, essential for analyses involving areas outside the original study locales. Lastly, to ensure data integrity, the \texttt{whichNA\_dropped} output identifies those observations that were excluded because of missing values. 

\begin{figure}[ht]
 \begin{center}  
\includegraphics[width=0.69\linewidth]{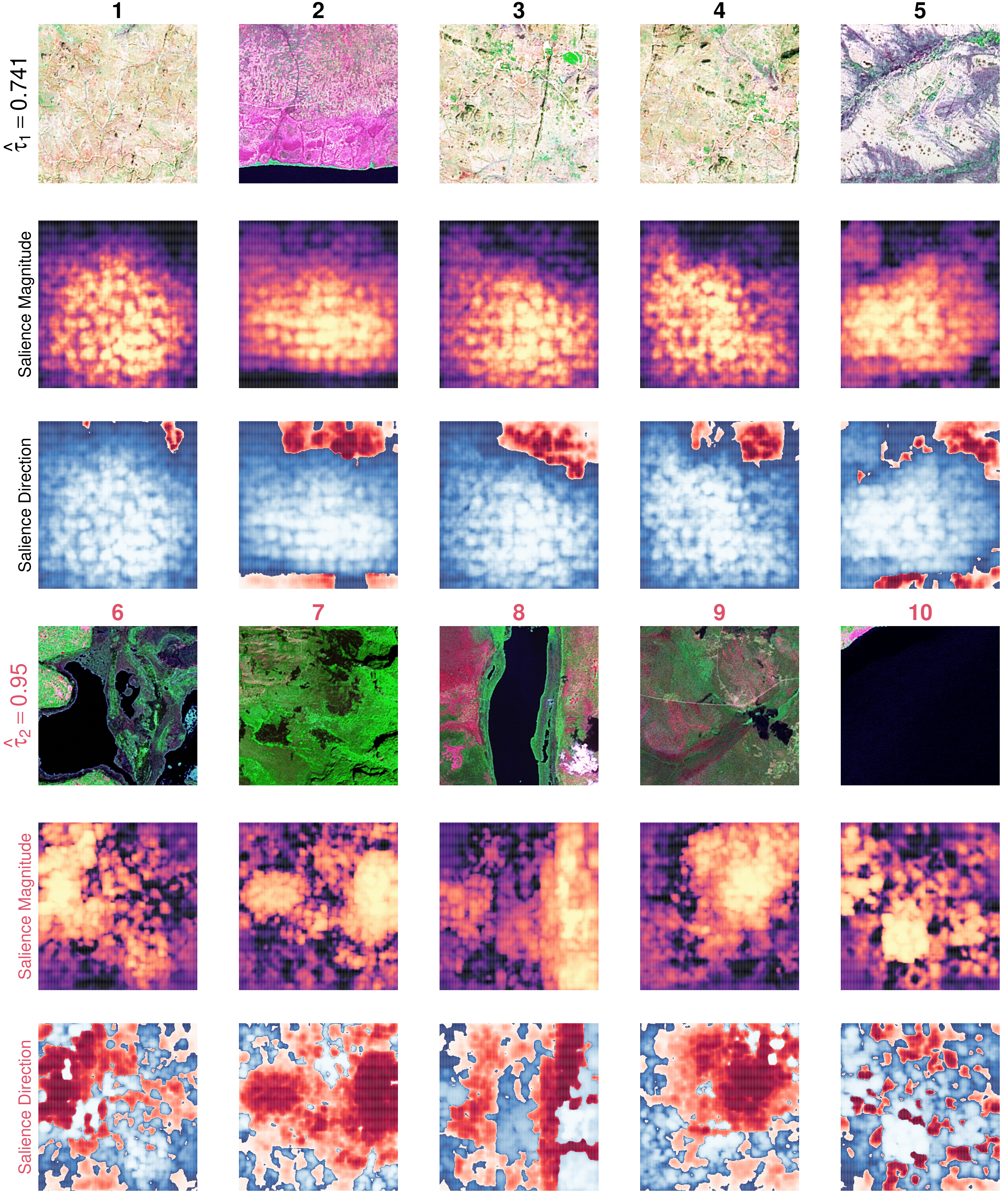}
\caption{\emph{Left, top 3 rows:}  High  probability cluster 1 images. \emph{Left, bottom 3 rows:} High probability cluster 2 images. ``Salience Magnitude'' and ``Direction'' represent two ways of analyzing salience. See \cite{jejoda2022_hetero} for details. 
}\label{fig:ImageExemplars}
\end{center}
\end{figure}

In addition to providing these estimated quantities, the function also writes to disk summary PDF outputs containing salience maps, which quantify areas in the image or image sequence that, if changed, would lead to the largest change in predicted treatment effect cluster probability. An example of such figures is found in Figure \ref{fig:ImageExemplars}. 

We note that the image (sequence) heterogeneity analysis can be made for either the embeddings or CNN image modeling backbone. 


\section{Conclusion and Future Development} \label{sec:summary}
As previously mentioned, {\bf causalimages} closes the gap between foundational and applied research in using image data for causal inference \citep{jerzak2022estimating,jejoda2022_hetero}. Besides the application cases discussed, we expect a take up in climate research, particularly pertaining to research in natural disaster evaluation \citep{shiba_heterogeneity_2021,shiba_estimating_2021,kakooei_chapter_2022,daoud_what_2016}, armed conflict, and ecology. These research fields include processes occurring on the surface of the earth with often substantial impact that is measurable from space. In the age of data science, an increasing number of researchers are using image data \citep{daoud_statistical_2023}. Although the package focuses on earth observation and global development research, the package is usable across a variety of domains in economics \citep{hall_remote_2010,henderson_measuring_2012}, sociology \citep{daoud_using_2023}, public policy \citep{balgiPersonalizedPublicPolicy2022}, public health \citep{kino_scoping_2021,conklin_impact_2018},  and biomedical applications \citep{castroCausalityMattersMedical2020}. 

Having discussed the current functionalities of {\bf causalimages}, we now turn to the future development of it.  We discuss four developments we will implement in the near future. First, we will modularize the image models that power {\bf causalimages}, allowing the user to plug in their preferred image model or extend it with a different backbone (i.e., feature extractor). Currently, the {\bf causalimages} uses two types of image-modeling backbones: the randomized embedding and the CNNs. While the CNN is a supervised procedure, the embedding backbone is unsupervised. However, there are many different foundational image-processing models or model architectures that the user might wish to consider using---VGGs, ResNets, U-NETs, LSTMS, Inception, and Transformers-based models. 

These models allow the user to calibrate their modeling approach to the data at hand. Additionally, there are several pre-trained models trained on classical datasets such as ImageNEt or CIFAR, or earth observation data. For example, recently, NASA and IBM trained such a model \citep{fraccaro_hls_2023}. Using these models in combination with the principles of transfer learning, will enable researchers to adapt their image models to often small datasets that we encounter in the biomedical and social sciences. Thus, modularization will enable the user to adapt {\bf causalimages} for their data and modeling needs.

Second, we will create image simulation facilities for causal inference. Often, when researchers develop or use methods in observational studies, they wish to simulate data to gain a deeper understanding of a causal system of interest. However, because images are high-dimensional objects, with a vast amount of parameters, it can be challenging to simulate image data that associates in a desired way with tabular data. To enable such simulations, we will incorporate a set of generative models to simulate counterfactual image scenarios. These generative models will partly build on the deep geography literature \citep{zhao_deep_2021}.

Third, we will incorporate existing models for causal discovery. There is considerable literature on discovery in high-dimensional data (cite Bernard group), and by connecting to that research, we foresee a cross-fertilization between causal inference conduct in a deductive versus inductive manner. Thus, we will incorporate those that are able to detect signals in image data \citep{lopez-paz_discovering_2017}. 

Fourth, we will incorporate different uncertainty quantification. Currently, {\bf causalimages} is able to quantify sampling uncertainty by using bootstrapping or a Bayesian approach. However, both approaches can be computationally expensive. Thus, a future area is to improve computational efficiency and keep updating the package to incorporate insights from the state of art \citep{smith_uncertainty_2014,abdar_review_2021}.  

Fifth, we will likely need to develop a grammar for causal inference with image data--inspired by the vision of the grammar of graphics \citep{wilkinson_grammar_2005,tufte_visual_2001}. As we discuss in \citet{jerzak2022estimating,jejoda2022_hetero}, image data comes with varying bands, resolution, and revisiting time. Thus, they have varying data structures. That also implies that the extent to which image data provides a window to causally analyzing the phenomena of interest will vary with these data structures. To handle that variability, researchers will likely need to have different functions or arguments to work efficiently and precisely with these data. That grammar development entails that we need to both align our {\bf causalimages} with existing functions in common geospatial packages as well as develop extension software. 


\section*{Acknowledgments}
We thank the members of the AI and Global Development Lab: James Bailie, Cindy Conlin, Devdatt Dubhashi, Felipe Jordan, Mohammad Kakooei, Eagon Meng, Xiao-Li Meng, and Markus Pettersson for valuable feedback on this project. We also thank Xiaolong Yang. In particular, we would like to acknowledge Cindy Conlin for being the first user of the package and for providing excellent feedback.

\bibliography{causalimagesbib,refs,referencesAdel}

\newpage
\begin{appendix}
\section{Computational Details}\label{s:ComputationalDetails}
The results in this paper were obtained using
\proglang{causalimages}~0.0.1 with the
\pkg{tensorflow}~2.14.0 package. \proglang{R} itself
and all packages used are available from the Comprehensive
\proglang{R} Archive Network (CRAN) at
\url{https://CRAN.R-project.org/}.

\end{appendix}



\end{document}